\documentclass[preprint,12pt,sort&compress]{elsarticle}

\usepackage{textcomp,booktabs}
\usepackage[usenames,dvipsnames]{color}
\usepackage{colortbl}
\definecolor{mygray}{gray}{.9}
\definecolor{mypink}{rgb}{.99,.91,.95}
\definecolor{mycyan}{cmyk}{.3,0,0,0}
\usepackage{amssymb}
\usepackage{graphicx}
\usepackage{booktabs}
\usepackage{tabularx}
\usepackage{array}
\usepackage{lscape}
\usepackage{longtable}
\usepackage{multirow}
\usepackage{amsmath}
\usepackage{colortbl}
\usepackage[figuresright]{rotating}
\usepackage{epsfig}
\usepackage{epsf}
\usepackage{subfigure}

\newcommand{\PreserveBackslash}[1]{\let\temp=\\#1\let\\=\temp}
\newcolumntype{C}[1]{>{\PreserveBackslash\centering}p{#1}}
\newcolumntype{R}[1]{>{\PreserveBackslash\raggedleft}p{#1}}
\newcolumntype{L}[1]{>{\PreserveBackslash\raggedright}p{#1}}
\newtheorem{definition}{Definition}[section]

\journal{} 
\begin{document}

\begin{frontmatter}



\title{Evidential supplier selection based on interval data fusion}


\author[address1]{Zichang He}
\author[address1]{Wen Jiang\corref{label1}}
\address[address1]{School of Electronics and Information, Northwestern Polytechnical University, Xi'an, Shaanxi, 710072, China}
\cortext[label1]{Corresponding author at: School of Electronics and Information, Northwestern  Polytechnical University, Xi'an, Shaanxi 710072, China. Tel: (86-29)88431267. E-mail address: jiangwen@nwpu.edu.cn, jiangwenpaper@hotmail.com}

\begin{abstract}
Supplier selection is a typical multi-criteria decision making(MCDM) problem and lots of uncertain information exist inevitably. To address this issue, a new method was proposed based on interval data fusion. Our method follows the original way to generate classical basic probability assignment(BPA) determined by the distance among the evidences. However, the weights of criteria are kept as interval numbers to generate interval BPAs and do the fusion of interval BPAs. Finally, the order is ranked and the decision is made according to the obtained interval BPAs. In this paper, a numerical example of supplier selection is applied to verify the feasibility and validity of our method. The new method is presented aiming at solving multiple-criteria decision-making problems in which the weights of criteria or experts are described in fuzzy data like linguistic terms or interval data.
\end{abstract}


\end{frontmatter}
\section{Introduction}\label{Introduction}
%\begin{table}[t]
%\caption{Detailed 20 individuals of each population excluding the
%outgroup (20) Table}\label{tbl1}
%\begin{tabular*}{\textwidth}{@{\extracolsep\fill}llll}
% \toprule
% A& Species\\ \cmidrule{2-4}
% & {\it T. acteon} & {\it T. sylvestris} & {\it T. lineola}\\
%\midrule variance among populations (all) & 0.0718 & 0.0179 &
%0.0081\\ $F_{ST}$ & 5.1\% & 1.6\% & 0.9\%\\ $p$ & <0.0001 &
%<0.0001 & 0.21\\
% variance among populations (20) & 0.0755 & 0.0256 &
%0.0077\\
% $F_{ST}$ & 5.3\% & 2.3\% & 0.8\%\\
%p & <0.0001 & <0.011 & 0.86\\
% \bottomrule
%\end{tabular*}
%\end{table}

Multiple-criteria decision-making has wide application in dealing with the comparison of multiple decisions. Because many decision-making projects like supplier selection will inevitably include the consideration of evidence based on several criteria\cite{choy2005knowledge}, rather than on a preferred single criterion in real world. An effective framework for decisions comparison based on the evaluation of multiple criteria is proposed in MCDM. It means that an optimal decision will be made based on comprehensive consideration. Compared with those approaches based on experience and intuition, MCDM is apparently more objective and reasonable\cite{mateos2014dominance,Kontio1996A,hung2012online}.

In realistic situation, selections often proceed under the environment occupied with unknown and uncertain information. As the the complexity of the system grows, the uncertainty of the problems and the fuzziness of human's thinking constantly increase accordingly. Hence, It is difficult for people to judge and distribute the importance of each criterion in MCDM. Fuzzy sets theory introduced by Zadeh is a good approach to settle with the uncertain information\cite{Jiang2015improved,jiang2015determining,Chou2016A}. And it has wide application in MCDM\cite{Liang1999Fuzzy,Wu2009Supplier,Hu2009Fuzzy,Deng2006Plant}. Decision making and optimization under uncertain environment is heavily studied\cite{Zhang2015Supplier,Jiang2016A}. Besides fuzzy sets theory, D-S evidence theory introduced by Dempster and Shafe\cite{Dempster1967Upper,Shafer1978A} plays an important role in making decisions under uncertain environment\cite{Jiang2016AN,fu2015group,Su2015Combining}. Due to the efficiency modeling and fusion of information, evidence theory is widely used\cite{deng2015Generalized,Jiang2016AF,Han2015Evaluations}. D-S evidence theory is also a powerful tool to deal with MCDM problems\cite{wang2016combination,AnImprovedAPIN2017,Dezert2016Decision,Mercier2009Decision,Yang2016new}.

As an effective tool to deal with the fuzzy data, interval number has been widely applied in MCDM. Chen and Chen-Tung(2000) defined a preference relation between each pair of plant locations based on the interval analysis and proposed  ranking method to determine the ranking order of all candidate location\cite{chen2000fuzzy}. Some paper like Janhanshanloo et.al(2006)\cite{Jahanshahloo2006An}, Yue and Zhongliang(2011)\cite{Yue2011A} and Liu et.al(2013)\cite{Liu2013New} proposed their methods to generate weights of criteria based on interval data. And Deng et.al(2011)\cite{Deng2011A} converted interval number to a crisp weight based on distance function and proposed a method to combine D-S evidence theory and fuzzy set theory to address MCDM problems.

In this paper, we propose a new method to solve MCDM problems based on the interval data fusion. In our method, interval data is retained during fusion process, which has some appreciable properties. First, interval data reflects the concrete and detailed information of the objects in a great extent. Additionally, it has some practical applications in some specific situations. For example, when the system requires the extremely high degree of accuracy we can only employ the lower limiting value and abandon the rest information to make decisions. Second, the weights of criteria or the information sources are allowed to be modeled as fuzzy number. Because we can convert the fuzzy description into interval data based on fuzzy set theory(FST). This property is quite useful in that not only quantitative data but also the qualitative representation is widely used in the practical decision-making problems. Third, fusing evidence based on interval data conforms with the universal cognition. Crisp data is a special form of interval data(like 0.5 can be seen as [0.5, 0.5]) in a way. In other words, our method is a generalized one of Deng et.al(2011)\cite{Deng2011A}.

The rest of this paper is organized as follows. The preliminaries of the basic theory employed are briefly presented in Section 2. And then our new fusion method based on interval data is proposed in Section 3. Section 4 takes a numerical example of supplier selection to show the efficiency of the method. Finally, the paper is concluded in Section 5.
\section{Preliminaries}
In this section, some preliminaries such as interval number, fuzzy set theory(FST), Dempster-Shafer theory(DST) and Pignistic probability transformation(PPT) are briefly introduced.

\subsection{Interval number}
\begin{definition}(Interval number) An interval number ${\tilde a}$ is defined as $\tilde a = [{a^L},{a^U}] = \{ x|{a^L} \le x \le {a^U}\} $ where ${a^L}$ is the lower limiting value and ${a^U}$ is the upper limiting value
while ${\rm{x}} \in \left[ {0,1} \right]$.
\end{definition}
Let ${\tilde a}$ and ${\tilde b}$ be two arbitrary positive closed interval numbers. The basic algorithm of interval number is given as follows\cite{Bosc1997An}:

\begin{equation}{
\tilde a + \tilde b = [{a^L} + {b^L},{a^U} + {b^U}]}
\end{equation}

\begin{equation}{
\tilde a \times \tilde b = [{a^L}{b^L},{a^U}{b^U}]}
\end{equation}

\begin{equation}{
\tilde a \div \tilde b = [\frac{{{a^L}}}{{{b^L}}},\frac{{{a^U}}}{{{b^U}}}]
}
\end{equation}

\begin{equation}{
k\tilde a = [k{a^L},k{a^U}]}
\end{equation}

\begin{equation}{
\frac{1}{{\tilde a}} = [\frac{1}{{{a^U}}},\frac{1}{{{a^L}}}]}
\end{equation}
for ${\tilde a}$ and ${\tilde b}$, let norm
$\left\| {\tilde a - \tilde b} \right\| = \left| {{a^L} - {b^L}} \right| + \left| {{a^U} - {b^U}} \right|$ be the so-called distance between the interval number ${\tilde a}$ and ${\tilde b}$.
Apparently, the larger $\left\| {\tilde a - \tilde b} \right\|$ is, the more ${\tilde a}$ and ${\tilde b}$ differ. Especially, interval number ${\tilde a}$ equals to ${\tilde b}$ completely when $\left\| {\tilde a - \tilde b} \right\| = 0$.
\subsection{Fuzzy sets theory}
Fuzzy set Introduced by Zadeh is an extension of classic set\cite{Zadeh1965Fuzzy}. It is an efficient tool to model linguistic variables.
\subsubsection{Fuzzy number}
A fuzzy set is any set that allows its members to have different grades of membership in the interval [0,1]. It consists of two components: a set and a membership function associated with it.
\begin{definition}(Fuzzy set). Let X be a collection of objects denoted generally by x, a fuzzy subset of X ${\tilde A}$ is a set of ordered pairs\cite{Gupta1992Fuzzy}:
\begin{equation} {
\tilde A = \left\{ {\left( {x,{\mu _{\tilde A}}\left( x \right)|x \in X} \right)} \right\}}
\end{equation}
${{\mu _{\tilde A}}\left( x \right)}$ is called the membership function (generalized characteristic function) which maps X to the membership space M. Its range is the subset of nonnegative real members whose supreme is finite.
\end{definition}

\begin{figure}[!ht]
\centering
\includegraphics[scale=0.5]{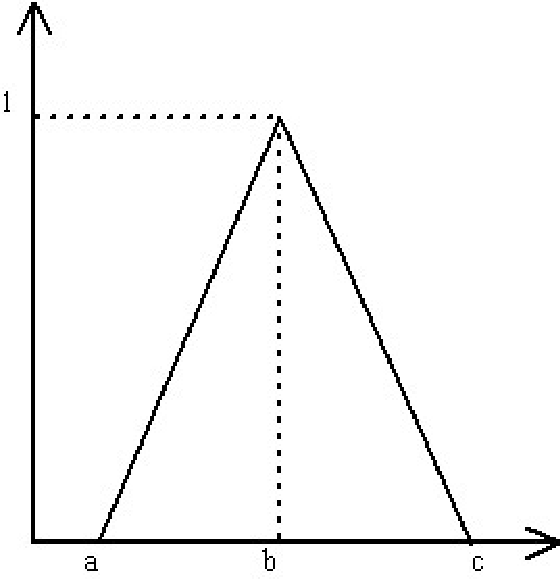}
 \caption{A triangular fuzzy number}\label{sanjiao}
\end{figure}

\begin{definition}(Triangular fuzzy number). A fuzzy number is a fuzzy subset of X. And a triangular fuzzy number ${\tilde A}$ can be defined by a triplet (a,b,c) shown in Fig. \ref{sanjiao}. Its membership function is defined as\cite{Zimmermann1991Fuzzy}
\begin{equation}{
{\mu _{\tilde A}}\left( x \right) = \left\{ {\begin{array}{*{20}{l}}
{\begin{array}{*{20}{c}}
{{\kern 5pt} {\kern 1pt} {\kern 1pt} {\kern 1pt} {\kern 1pt} {\kern 1pt} {\kern 1pt} {\kern 1pt} {\kern 1pt} 0,}&{x < a}
\end{array}}\\
{\begin{array}{*{20}{c}}
{{\textstyle{{x - a} \over {b - a}}},}&{a \le x \le b}
\end{array}}\\
{\begin{array}{*{20}{c}}
{{\textstyle{{c - x} \over {c - b}}},}&{b \le x \le c}
\end{array}}\\
{\begin{array}{*{20}{c}}
{{\kern 5pt} {\kern 1pt} {\kern 1pt} {\kern 1pt} {\kern 1pt} {\kern 1pt} {\kern 1pt} {\kern 1pt} 0,}&{x > c}
\end{array}}
\end{array}} \right.
}\end{equation}
\end{definition}

\subsubsection{Linguistic variable}
Linguistic variable is a variable with linguistic words or sentences in a natural language\cite{Zadeh1975The}. It is widely used in practical life and it is one of the most classical fuzzy information. When dealing with situations which are too complex or ill-defined to be accurately described in conventional quantitative expressions, it's convenient and reasonable to do a qualitative description. Generally, each linguistic variable corresponds to a fuzzy set. For example, these linguistic variables can be expressed in positive triangular fuzzy number\cite{Kaufmann1985Introduction} as Table 1.
\begin{table}[!h]
{\footnotesize\label{table2}
\caption{Linguistic variables for importance in triangular fuzzy number\cite{Kaufmann1985Introduction}}
\begin{tabular*}{\columnwidth}{@{\extracolsep{\fill}}@{~~}cccc@{~~}}
\toprule
  Terms            & triangular fuzzy number  \\
\midrule
  Very low (VL)    & (0, 0.1, 0.3)         \\
  Low (L)          & (0.1, 0.3, 0.5)       \\
  Medium (M)       & (0.3, 0.5, 0.7)       \\
  High (H)         & (0.5, 0.7, 0.9)       \\
  Very high (VH)   & (0.7, 0.9, 1.0)       \\
\bottomrule
\end{tabular*}
}
\end{table}
\\
Virtually, the concrete models used to represent the linguistic items are flexible and changeable. To apply which kind of represent method depends on the realistic application systems and the domain experts' opinions. In MCDM problems like supplier selection in Section 4, our method adopted the method which converts the linguistic variable into interval data.

\subsection{Dempster-Shafer theory of evidence}
Dempster-Shafer theory is a mathematical theory of evidence which is used to combine separate pieces of information(evidence) to calculate the belief probability of an event.
In a D-S theory reasoning scheme, the set of possible hypotheses are collectively called the frame of discernment $\Theta $, defined as follows\cite{Shafer1978A}:
 \[\Theta {\rm{ = }}\left\{ {{H_1},{H_2},{H_3}, \ldots  \ldots ,{H_{\rm{n}}} } \right\}\]
where n is the number of exclusive and exhaustive elements in the set. Form the frame of discernment $\Theta $, let $P\left( \theta  \right)$ denote the power set composed with the ${2^N}$ propositions A of $\Theta $:
\[P\left( \theta  \right) = \left\{ {\emptyset ,\left\{ {{H_1}} \right\},\left\{ {{H_2}} \right\}, \ldots  \ldots \left\{ {{H_{\rm{n}}}} \right\},\left\{ {{H_1} \cup {H_2}} \right\},\left\{ {{H_1} \cup {H_3}} \right\}, \ldots  \ldots \theta } \right\}\]
where $\emptyset$ denotes the empty set. Then a mass function $m$ is defined as
${\rm{m}}\left( {{2^\theta }} \right) \in \left[ {0,1} \right]$ to distribute the belief across the frame meeting the following conditions:

\begin{center}{$m\left( \emptyset  \right) = 0$ and $\sum\nolimits_{A \subseteq \theta } {m\left( A \right)}  = 1$}
\end{center}

Under these circumstances, the beliefs of the evidence source can only be assigned to non-empty hypotheses and must sum to 1. When the belief is assigned to one hypothesis, the more elements the hypothesis contains, the less information it offers. Especially, a hypothesis containing all the elements means nothing is informative essentially. For the algorithm designed to access evidence, the most significant ability is to combine the evidence from multiple sources. And the crucial process of combining two pieces of evidence from independent sources is fulfilled with the following equation called Dempster's combination rule:
\begin{equation}\label{Eq8}{
{m_{12}}\left( A \right) = \frac{{\sum\nolimits_{\forall x,y:x \cap y = A} {{m_1}\left( X \right) \cdot {m_2}\left( Y \right)} }}{{1 - K}}
}\end{equation}
with
\begin{equation}{
K{\rm{ = }}\sum\nolimits_{\forall X,Y:X \cap Y = \emptyset } {{m_1}\left( X \right) \cdot {m_2}(Y)} }
\end{equation}
where ${m_{12}}\left( A \right)$ is the new belief for the hypothesis $A$ yielded from the original evidence ${m_1}$ and ${m_2}$. Apparently, Eq.(\ref{Eq8}) can only be applied when $K \ne 0$. $K$ is called the conflict coefficient and $1{\rm{ - }}K$ is a constant coefficient used to normalize the combined evidence. And all combined evidence whose intersection is not the hypothesis of interest $A$ is represented by K. Its value reveals the degree of the confliction between the two original evidence, $K = 0$ means the consistence of the belief assignment, whereas $K = 1$ means the complete contradiction .
\\Likewise, when the evidence is from $j$ different sources, the rule can be expressed as:
\begin{equation}{
m\left( A \right) = \frac{{\sum\nolimits_{\forall X,Y \ldots Z{\kern 1pt} {\kern 1pt} :{\kern 1pt} {\kern 1pt} {\kern 1pt} X \cap Y \ldots  \cap Z = A} {{m_1}\left( X \right) \cdot {m_2}\left( Y \right) \cdots {{\rm{m}}_j}(Z)} }}{{1 - K}}
}\end{equation}
with
\begin{equation}{
K{\rm{ = }}\sum\nolimits_{\forall X,Y \ldots Z{\kern 1pt} {\kern 1pt} :{\kern 1pt} {\kern 1pt} {\kern 1pt} X \cap Y \ldots  \cap Z = \emptyset } {{m_1}\left( X \right) \cdot {m_2}\left( Y \right) \cdots {{\rm{m}}_j}(Z)} {\rm{ }}
}\end{equation}

\subsection{Pignistic probability transformation}
Virtually, two levels are classified to describe the beliefs: one is the credal level where belief is entertained. And the other one is the pignistic level where beliefs are feasible to make decisions\cite{Smets1994The}. The term "pignistic" proposed by Smets is originated from the word pignus, meaning 'bet' in Latin. Pignistic probability has a wide application on decision-making. Principle of insufficient reason is used to assign the basic probability of multiple-element set to singleton set. In other word, a belief interval is distributed into the crisp ones determined as:
\begin{equation}\label{Eq12}{
bet\left( {{A_i}} \right) = \sum\nolimits_{{A_i} \subseteq {A_k}} {\frac{{m\left( {{A_k}} \right)}}{{\left| {{A_k}} \right|}}}
}\end{equation}
where ${\left| {{A_k}} \right|}$ denotes the number of elements in the set called the cardinality. Eq.(\ref{Eq12}) is also called as Pignistic Probability Transformation(PPT).

\section{Proposed method}\label{Proposed method}
In this section, our new method based on interval data fusion is proposed. In general, a basic MCDM problem can be modeled as follows:
For a certain problem, there is a committee of k decision-makers $\left\{ {{D_1},{D_2},{D_3}, \ldots  \ldots ,{D_k}} \right\}$ to evaluate it. Each decision maker holds m alternatives  $\left\{{{A_1},{A_2},{A_3}, \ldots  \ldots ,{A_m}} \right\}$. And for each alternative, n criteria $\left\{ {{C_1},{C_2},{C_3}, \ldots  \ldots ,{C_n}} \right\}$ are in consideration to make decisions(usually  the same criterion is shared). The following is a succinct model proposed by Hwang and Yoon\cite{Hwang1981Multiple} to express MCDM in a matrix format.
\[\begin{array}{l}
{\kern 1pt} {\kern 1pt} {\kern 1pt} {\kern 1pt} {\kern 1pt} {\kern 1pt} {\kern 1pt} {\kern 1pt} {\kern 1pt} {\kern 1pt} {\kern 1pt} {\kern 1pt} {\kern 1pt} {\kern 1pt} {\kern 1pt} {\kern 1pt} {\kern 1pt} {\kern 1pt} {\kern 1pt} {\kern 1pt} {\kern 1pt} {\kern 1pt} {\kern 1pt} {\kern 1pt} {\kern 1pt} {\kern 1pt} {\kern 1pt} {\kern 1pt} {\kern 1pt} {\kern 1pt} {\kern 1pt} {\kern 1pt} {\kern 1pt} {\kern 1pt} {\kern 1pt} {\kern 1pt} {\kern 1pt} {\kern 1pt} {\kern 1pt} {\kern 1pt} {\kern 1pt} {\kern 1pt} {\kern 1pt} {\kern 1pt} {\kern 1pt} {\kern 1pt} {\kern 1pt} {\kern 1pt} {\kern 1pt} {\kern 1pt} {\kern 1pt} {\kern 1pt} {\kern 1pt} {\kern 1pt} {\kern 1pt} {\kern 1pt} {\kern 1pt} {\kern 1pt} {\kern 1pt} {\kern 1pt} {\kern 1pt} {\kern 1pt} {\kern 1pt} {\kern 1pt} {\kern 1pt} {\kern 1pt} {\kern 1pt} {\kern 1pt} {\kern 1pt} {\kern 1pt} {\kern 1pt} {\kern 1pt} {\kern 1pt} {\kern 1pt} {\kern 1pt} {\kern 1pt} {\kern 1pt} {\kern 1pt} {\kern 1pt} {\kern 1pt} {\kern 1pt} {\kern 1pt} {\kern 1pt} {\kern 1pt} {\kern 1pt} {\kern 1pt} {\kern 4pt} {\begin{array}{*{20}{c}}
{{C_1}}&{{C_2}}& \ldots &{{C_n}}
\end{array}}{\kern 1pt} {\kern 1pt} {\kern 1pt} {\kern 1pt} {\kern 1pt} {\kern 1pt} {\kern 1pt} {\kern 1pt} \\
{D_k} = {\kern 1pt} {\kern 1pt} {\kern 1pt} {\kern 1pt} {\kern 1pt} \begin{array}{*{20}{c}}
{{A_1}}\\
{{A_2}}\\
 \vdots \\
{{A_m}}
\end{array}{\kern 1pt} {\kern 1pt} {\kern 1pt} {\kern 1pt} {\kern 1pt} {\kern 1pt} {\kern 1pt} {\kern 1pt} {\kern 1pt} \left[ {\begin{array}{*{20}{c}}
{{r_{11}}}&{{r_{12}}}& \ldots &{{r_{1n}}}\\
{{r_{21}}}&{{r_{22}}}& \ldots &{{r_{2n}}}\\
 \vdots & \vdots & \ldots & \vdots \\
{{r_{m1}}}&{{r_{m2}}}& \ldots &{{r_{mn}}}
\end{array}} \right]
\end{array}\]
where ${{r_{mn}}}$ is the rating of alternative ${{A_m}}$ with respect to criteria ${{C_n}}$ which is usually described crisply or fuzzily. In our method, ${{r_{mn}}}$ is allowed to be a crisp number or in the form of an interval data. For now, the facing problem is how to acquire ${{r_{mn}}}$.

In the practical, the final aim is often to rank the alternatives and make the best selection. Accordingly, the final scores of every alternative are not cared too much. Considering that, Hwang and Yoon proposed TOPSIS(Technique for Order Preference by Similarity to Ideal Solution) to solve MCDM\cite{Roh2000In}. The principle is that the chosen alternative should have the shortest distance from the positive ideal solution and the farthest distance from the positive ideal solution. Based on TOPSIS, Deng et al.\cite{Deng2011A} proposed a new method using FST together with DST. In that method, the ideal solution , negative ideal solution is determined and the distances of an alternative between them are determined. Then the classical BPA is generated to describe how close between both the alternative to ideal solution and to negative ideal solution.

In the classical TOPSIS, the performance ratings and the weights of the criteria are given as crisp values. Hence, Deng et al.\cite{Deng2011A} changed the fuzzy MCDM problem into a crisp one via using the distance function. However, his method only average the lower limitation and the upper limitation of the interval. The new crisp weights are generated according to the average in the essence. It means that one criterion holds the weight of $[0.1,0.9]$ measures the same as another one holds $[0.4,0.6]$, which is apparently not reasonable and convincing enough. Because for one criterion with a weight in the form of an interval number ${\tilde a}$, a smaller result of ${a^U} - {a^L}$ represents that the information about the criterion is more clear when the sum of ${a^L}$ and ${a^U}$ is constant. Whereas the above, it's rational for us to allocate a larger crisp weight to the criterion which weighs $[0.4,0.6]$ than the one holding $[0.1,0.9]$ .
To improve Deng et al.'s method\cite{Deng2011A}, we retain the interval data in the fusion procedure and generate the interval BPA. Based on the TOPSIS, the elements of our interval BPA are \{IS(ideal solution)\},  \{NS(negative solution)\} and \{IS,NS\}, of which {IS,NS} is the frame of discernment. The following is the example of the interval BPA for one certain alternative£º
\[{m_1}(\left\{ {IS} \right\}) = [{a^L},{a^U}]\]
\[{m_1}(\left\{ {NS} \right\}) = [{b^L},{b^U}]\]
\[{m_1}(\left\{ {IS,NS} \right\}) = [{c^L},{c^U}]\]

It means that:

1) The hypothesis "the alternative is an ideal solution"  is upheld with belief degree from ${a^L}$ to ${a^U}$.

2) The hypothesis "the alternative is a negative ideal solution"  is upheld with belief degree from ${b^L}$ to ${b^U}$.

3) The hypothesis "the alternative is perceived as a discernment, namely it is likely to be an ideal solution or a negative solution" is upheld with belief degree from ${c^L}$ to ${c^U}$.

It is worth mentioning that ${m_1}(\left\{ {IS,NS} \right\}) = 1 - {m_1}(\{ IS\})  - {m_1}(\{ NS\}) $. Hence, it's easy to know that ${c^L} = 1 - {a^U} - {b^U}$ and ${c^U} = 1 - {a^L} - {b^L}$.

For the focal element $(\{ IS\} ,\{ NS\} ,\{ IS,NS\} )$, there is another way to express interval BPA as $\{ [{a^L},{b^L},{c^L}],[{a^U},{b^U},{c^U}]\}$. When making a decision based on interval BPA, we can fuse the BPA consisting of the lower limitation and the one consisting of the upper limitation into a classical BPA by fusing the left part and right part with \ref{Eq8}. In the ultimate, PPT is used to compare the BPA of {IS}. In accordance with the notion mentioned above, our new method can be stated step by step as follows:

\textbf{Step 1.} Determine the ideal solution and negative ideal solution. And generate the classical BPA of each performance based on the distance between IS and NS.

\textbf{Step 2.} Convert the criteria's weights including crisp data(0.5 can be seen as [0.5,0.5]) and linguistic items into an interval number. And then discount the classical BPAs using the interval data to generate the interval BPA of each performance. Combine the interval BPAs of each criterion to get one comprehensive evaluation of an alternative.

\textbf{Step 3.} Convert the decision makers' weights including crisp data and linguistic items into interval numbers. And then discount the interval BPAs of combined performance (obtained in Step 2) using the interval data to generate the interval BPA of each performance. Combine the interval BPAs of all decision makers' to get the performance of each alternative.

\textbf{Step 4.} Combine the the left part and the right part of the interval BPAs to get the final performance of each alternative.

\textbf{Step 5.} Compare and rank the order of decisions based on PPT and make the best decision.

\section{Numerical example}\label{Numerical example}
 Supplier selection is a typical MCDM problem where lots of fuzzy information exist. In reality, although managers claims that the quality is the most important attribute for a supplier, they actually choose suppliers based largely on cost and delivery performance\cite{Verma1998An}. To identify the availability of our new method, the numerical example used in paper\cite{Deng2011A} is adopted in this section. The initial condition, such as the classical BPA of each performance and the weights of each criterion as well as the weights of experts are shown in Table \ref{Data for supplier selection(Deng,2011)}.
\begin{sidewaystable}
{
\caption{Data of supplier selection in Deng et.al(2011)}\label{Data for supplier selection(Deng,2011)}
\scriptsize
\begin{tabular*}{\columnwidth}{@{\extracolsep{\fill}}@{~~}llllll@{~~}}
\toprule
                 &Performance  & $C1$                         & $C2$                         & $C3$                        & $C4$                         \\
                 &             &$(\{IS\},\{NS\},\{IS,NS\})$   &$(\{IS\},\{NS\},\{IS,NS\})$   &$(\{IS\},\{NS\},\{IS,NS\})$  &$(\{IS\},\{NS\},\{IS,NS\})$   \\
\midrule
  DM1            &Weights      &[0.20,0.35]                       &[0.30,0.55]                       &[0.05,0.30]                      &[0.25,0.50]                   \\
  {[0.20,0.45]}  &Supplier1    &(0.60,0.20,0.20)                  &(0.6429,0.0714,0.2857)            &(0.60,0.20,0.20)                 &(0.60,0.20,0.20)              \\
                 &Supplier2    &(0.60,0.20,0.20)                  &(0.6429,0.0714,0.2857             &(0.50,0.50,0)                    &(0.50,0.50,0)                 \\
                 &Supplier3    &(0.50,0.50,0)                     &(0.50,0.50,0)                     &(0.60,0.20,0.20)                 &(0.6667,0,0.3333)             \\
                 &Supplier4    &(0.66667,0,0.3333)                &(0.6667,0,0.3333)                 &(0.50,0.50,0)                    &(0.50,0.50,0)                 \\
                 &Supplier5    &(0,0.6667,0.3333)                 &(0,0.6667,0.3333)                 &(0,0.6667,0.3333)                &(0,0.6667,0.3333)             \\
                 &Supplier6    &(0.20,0.60,0.20)                  &(0.0714,0.6429,0.2857)            &(0.6667,0,0.3333)                &(0,0.6667,0.3333)             \\
                                                                                                                                                               \\
                                                                                                                                                               \\
  DM2            &Weights      &[0.25,0.45]                       &[0.20,0.55]                       &[0.05,0.3]                       &[0.20,0.60]                     \\
  {[0.35,0.55]}  &Supplier1    &(0.60,0.20,0.20)                  &(0.6429,0.0714,0.2857)            &(0.50,0.50,0)                    &(0.60,0.20,0.20)              \\
                 &Supplier2    &(0.60,0.20,0.20)                  &(0.6429,0.0714,0.2857)            &(0,0.6667,0.3333)                &(0.50,0.50,0)                 \\
                 &Supplier3    &(0.50,0.50,0)                     &(0.50,0.50,0)                     &(0.6667,0,0.3333)                &(0.6667,0,0.3333)             \\
                 &Supplier4    &(0.66667,0,0.3333)                &(0.6667,0,0.3333)                 &(0.50,0.50,0)                    &(0.50,0.50,0)                 \\
                 &Supplier5    &(0,0.6667,0.3333)                 &(0,0.6667,0.3333)                 &(0,0.6667,0.3333)                &(0.20,0.60,0.20)              \\
                 &Supplier6    &(0.20,0.60,0.20)                  &(0.0714,0.6429,0.2857)            &(0.6667,0,0.3333)                &(0,0.6667,0.3333)             \\
                                                                                                                                                               \\
                                                                                                                                                               \\
  DM3            &Weights      &[0.20,0.55]                       &[0.20,0.70]                       &[0.10,0.40]                      &[0.20,0.60]                   \\
  {[0.70,0.95]}  &Supplier1    &(0.60,0.20,0.20)                  &(0.6429,0.0714,0.2857)            &(0.5714,0.2857,0.1429)           &(0.6667,0,0.3333)             \\
                 &Supplier2    &(0.60,0.20,0.20)                  &(0.6429,0.0714,0.2857             &(0.6667,0,0.3333)                &(0.2857,0.5714,0.1429)        \\
                 &Supplier3    &(0.50,0.50,0)                     &(0.50,0.50,0)                     &(0.6667,0,0.3333)                &(0.6667,0,0.3333)             \\
                 &Supplier4    &(0.66667,0,0.3333)                &(0.6667,0,0.3333)                 &(0.5714,0.2857,0.1429)           &(0.6667,0,0.3333)             \\
                 &Supplier5    &(0,0.6667,0.3333)                 &(0,0.6667,0.3333)                 &(0,0.6667,0.3333)                &(0.2857,0.5714,0.1429)        \\
                 &Supplier6    &(0.20,0.60,0.20)                  &(0.0714,0.6429,0.2857)            &(0.6667,0,0.3333)                &(0,0.6667,0.3333)             \\
\bottomrule
\end{tabular*}
}
\end{sidewaystable}

As shown in Table \ref{Data for supplier selection(Deng,2011)}, there are four criteria, including product late delivery, cost, risk factor and suppliers' service performance detailed as following:

C1: Product late delivery. The delivery process can reflect the service ability of a supplier. It is considered to investigate whether the supplier can supply stable and constant appreciation serve for the enterprise.

C2: Cost. A good price measures quite a lot in reducing cost and increasing the competitive force.

C3: Risk factor. If we want to make long-term cooperation with a supplier, then we must take its risk factor (political factor, economic factor, the reputation, etc.) into account.

C4: Supplier's service performance. Service performance means the sustaining promotion of the product and service(e.g. product quality acceptance level, technological support, information process), which is deemed as the core factor.

It should be noticed that the weights are ready interval data. If they are described in the fuzzy linguistic items, we can also convert them into interval data. Table \ref{Linguistic variables for the importance weight and ratings} is an example in which the linguistic items and their according interval data differ in different situations. It is one of the remarkable advantages of our method. And the criterion of the value of interval number depends on the experts' opinions.
\begin{table}[!h]
{\footnotesize
\caption{Convert linguistic variables into the interval data}\label{Linguistic variables for the importance weight and ratings}
\begin{tabular*}{\columnwidth}{@{\extracolsep{\fill}}@{~~}cccc@{~~}}
\toprule
  Terms            & Interval data  \\
\midrule
  Very low (VL)    & [0, 0.3]       \\
  Low (L)          & [0.1, 0.5]     \\
  Medium (M)       & [0.3, 0.7]     \\
  High (H)         & [0.5, 0.9]     \\
  Very high (VH)   & [0.7, 1.0]     \\
\bottomrule
\end{tabular*}
}
\end{table}

\begin{figure}[!ht]
\centering
\includegraphics[scale=1]{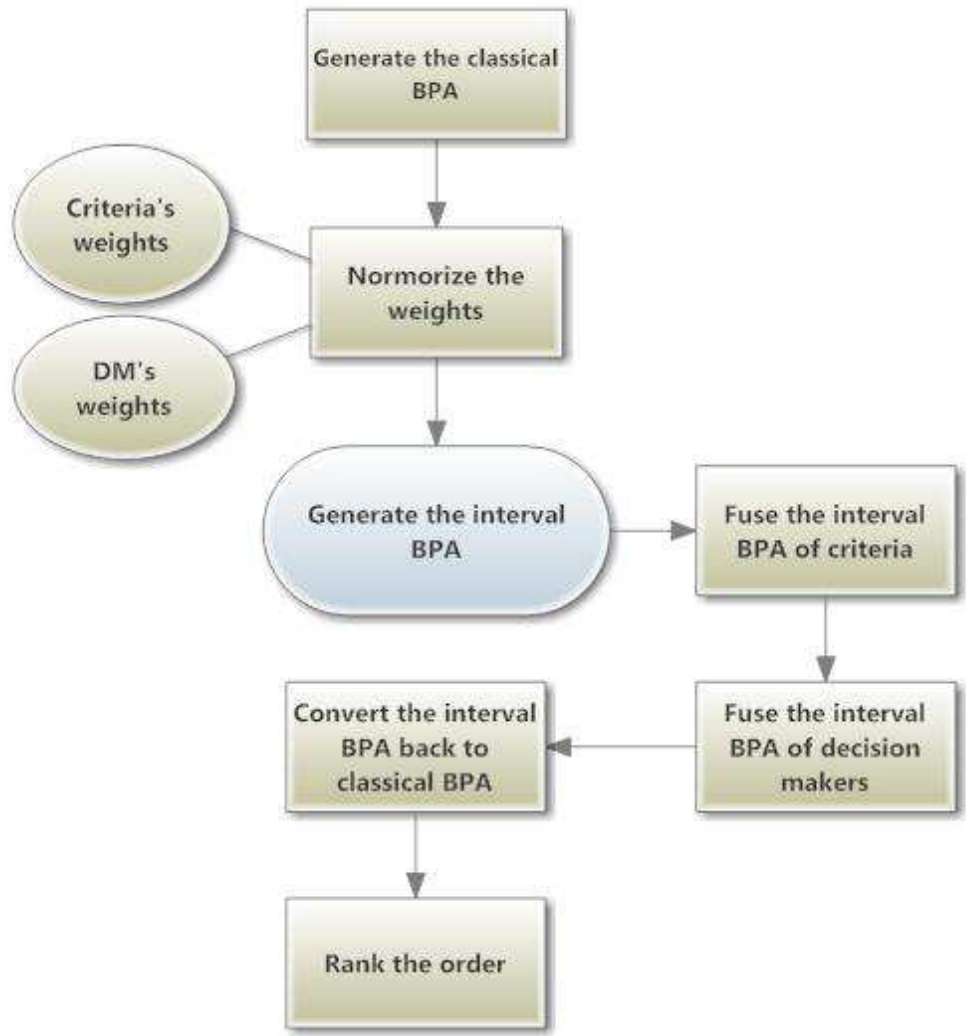}
 \caption{Supplier selection based on interval data fusion}\label{liuchengtu}
\end{figure}

Before applying our method, a flow chart(Figure \ref{liuchengtu}) is shown to summarize the whole procedure of applying our method in the supplier selection problem. Based on it, the detailed processes will be illustrated step by step in the following.
\\
\textbf{Step 1.} Determine the ideal solution and negative ideal solution. And generate the classical BPA of each performance based on the distance between IS and NS.

Since the classical BPA of performance is already known in Table 2, we will implement the following steps of our method to these data in order.

\textbf{Step 2.} Convert the criteria's weights including crisp data(0.5 can be seen as [0.5,0.5]) and linguistic items into an interval number. And then discount the classical BPA using the interval data to generate the interval BPA of each performance. Combine the interval BPA of each criterion to get one comprehensive evaluation of an alternative.

In this situation, after simple data processing we can put the weights of criteria into use directly, since all the original weights are in the form of interval data. What we only need to do is to normalize the interval number with the following equation:
\begin{equation}\label{Eq13}
{W = {{\left[ {\begin{array}{*{20}{c}}
{{a^L},}&{{a^U}}
\end{array}} \right]} \mathord{\left/
 {\vphantom {{\left[ {\begin{array}{*{20}{c}}
{{a^L},}&{{a^U}}
\end{array}} \right]} {{a_{\max }}}}} \right.
 \kern-\nulldelimiterspace} {{a_{\max }}}}}
\end{equation}

where ${{\rm{a}}_{\max }}$ is the largest number among all the limitation values of the intervals.

For example, the new weights of DM1's four criteria are as following respectively:
\[{W_{C1}} = {{[\begin{array}{*{20}{c}}
{0.20},&{0.35}
\end{array}]} \mathord{\left/
 {\vphantom {{[\begin{array}{*{20}{c}}
{0.20},&{0.35}
\end{array}]} {0.70}}} \right.
 \kern-\nulldelimiterspace} {0.70}} = \left[ {\begin{array}{*{20}{c}}
{0.2857},&{0.5}
\end{array}} \right]\]
\[{W_{C2}} = {{[\begin{array}{*{20}{c}}
{0.30},&{0.55}
\end{array}]} \mathord{\left/
 {\vphantom {{[\begin{array}{*{20}{c}}
{0.30},&{0.55}
\end{array}]} {0.70}}} \right.
 \kern-\nulldelimiterspace} {0.70}} = \left[ {\begin{array}{*{20}{c}}
{0.4286},&{0.7857}
\end{array}} \right]\]
\[{W_{C3}} = {{[\begin{array}{*{20}{c}}
{0.05},&{0.30}
\end{array}]} \mathord{\left/
 {\vphantom {{[\begin{array}{*{20}{c}}
{0.05},&{0.30}
\end{array}]} {0.70}}} \right.
 \kern-\nulldelimiterspace} {0.70}} = \left[ {\begin{array}{*{20}{c}}
{0.0714},&{0.4286}
\end{array}} \right]\]
\[{W_{C4}} = {{[\begin{array}{*{20}{c}}
{0.25},&{0.50}
\end{array}]} \mathord{\left/
 {\vphantom {{[\begin{array}{*{20}{c}}
{0.25},&{0.50}
\end{array}]} {0.70}}} \right.
 \kern-\nulldelimiterspace} {0.70}} = \left[ {\begin{array}{*{20}{c}}
{0.3571},&{0.7142}
\end{array}} \right]\]

let the new weight be $W = [{W_{\min }}, {W_{\max }}]$, then we can determine the interval BPA as
\begin{equation}
{m(\left\{ {IS} \right\}) = \left[ {{a^L}{W_{\min }},{\kern 1pt} {\kern 1pt} {\kern 1pt} {\kern 1pt} {\kern 1pt} {\kern 1pt} {\kern 1pt} {a^U}{W_{\max }}} \right]{\kern 1pt} {\kern 1pt} {\kern 1pt}}
\end{equation}

\begin{equation}
{m(\left\{ {NS} \right\}) = \left[ {{b^L}{W_{\min }},{\kern 1pt} {\kern 1pt} {\kern 1pt} {\kern 1pt} {\kern 1pt} {\kern 1pt} {\kern 1pt} {b^U}{W_{\max }}} \right]{\kern 1pt} {\kern 1pt} {\kern 1pt}}
\end{equation}

\begin{equation}
{m(\left\{ {IS,NS} \right\}) = \left[ {1 - {a^L}{W_{\min }} - {b^L}{W_{\min }}, 1 - {a^U}{W_{\max }} - {b^U}{W_{\max }}} \right]
}\end{equation}

So the integrated one is expressed as:
\begin{equation}\label{Eq17}{
\begin{array}{l}
m\left( {\{ IS\} ,\{ NS\} ,\{ IS,NS\} } \right) = \\
\left( {\begin{array}{*{20}{c}}
{\left[ {\begin{array}{*{20}{c}}
{{a^L}{W_{\min }},}&{{a^U}{W_{\max }}}
\end{array}} \right],}&{\left[ {\begin{array}{*{20}{c}}
{{b^L}{W_{\min }},}&{{b^U}{W_{\max }}}
\end{array}} \right]}
\end{array}} \right.,\\
\left. {{\kern 10pt} \left[ {\begin{array}{*{20}{c}}
{1 - {a^L}{W_{\min }} - {b^L}{W_{\min }},}&{1 - {a^U}{W_{\max }} - {b^U}{W_{\max }}}
\end{array}} \right]} \right)
\end{array}
}\end{equation}
or
\begin{equation}\label{Eq18}{
m{(\{ IS\} ,\{ NS\} ,\{ IS,NS\} ) = \left( {\begin{array}{*{20}{c}}
{{m_{left}},}&{{m_{right}}}
\end{array}} \right)}
}\end{equation}
with\\
${m_{left}} = \left[ {\begin{array}{*{20}{c}}
{{a^L}{W_{\min }},}&{{b^L}{W_{\min }},}&{1 - {a^L}{W_{\min }} - {b^L}{W_{\min }}}
\end{array}} \right]$ \\
${m_{right}} = \left[ {\begin{array}{*{20}{c}}
{{a^U}{W_{\max }},}&{{b^U}{W_{\max }},}&{1 - {a^U}{W_{\max }} - {b^U}{W_{\max }}}
\end{array}} \right]$

Let us take DM1's evaluation to C1 of supplier1 as an example:
\[{\rm{m}}_{C1}^1\left[ {IS} \right] = \left[ {\begin{array}{*{20}{c}}
{0.2875 \times 0.6},&{0.5 \times 0.6}
\end{array}} \right] = \left[ {\begin{array}{*{20}{c}}
{0.1714},&{0.3}
\end{array}} \right]\]
\[{\rm{m}}_{C1}^1\left[ {NS} \right] = \left[ {\begin{array}{*{20}{c}}
{0.2875 \times 0.2},&{0.5 \times 0.2}
\end{array}} \right] = \left[ {\begin{array}{*{20}{c}}
{0.0571},&{0.1}
\end{array}} \right]\]
\[\begin{array}{l}
{\rm{m}}_{C1}^1\left[ {IS,NS} \right] = \left[ {\begin{array}{*{20}{c}}
{1 - 0.2875 \times 0.6 - 0.2875 \times 0.2},&{1 - 0.5 \times 0.2 - 0.5 \times 0.6}
\end{array}} \right]\\
{\kern 1pt} {\kern 1pt} {\kern 1pt} {\kern 1pt} {\kern 1pt} {\kern 1pt} {\kern 1pt} {\kern 1pt} {\kern 1pt} {\kern 1pt} {\kern 1pt} {\kern 1pt} {\kern 1pt} {\kern 1pt} {\kern 1pt} {\kern 1pt} {\kern 1pt} {\kern 1pt} {\kern 1pt} {\kern 1pt} {\kern 1pt} {\kern 1pt} {\kern 1pt} {\kern 1pt} {\kern 1pt} {\kern 1pt} {\kern 1pt} {\kern 1pt} {\kern 1pt} {\kern 1pt} {\kern 1pt} {\kern 1pt} {\kern 1pt} {\kern 1pt} {\kern 1pt} {\kern 1pt} {\kern 1pt} {\kern 1pt} {\kern 1pt} {\kern 1pt} {\kern 1pt} {\kern 1pt} {\kern 1pt} {\kern 1pt} {\kern 1pt} {\kern 1pt} {\kern 1pt} {\kern 1pt} {\kern 1pt} {\kern 1pt} {\kern 1pt} {\kern 1pt} {\kern 1pt} {\kern 1pt} {\kern 1pt} {\kern 1pt} {\kern 1pt} {\kern 1pt} {\kern 1pt} {\kern 1pt} {\kern 1pt} {\kern 1pt} {\kern 1pt} {\kern 1pt} {\kern 1pt} {\kern 1pt} {\kern 1pt}  = \left[ {\begin{array}{*{20}{c}}
{0.7715},&{0.6}
\end{array}} \right]
\end{array}\]
By using the Eq.(\ref{Eq17}), the rest interval BPA of each performance is listed in Table \ref{Generating the interval BPAs of four criteria respectively}.

\begin{sidewaystable}
{
\caption{Generating the interval BPAs of four criteria respectively.}\label{Generating the interval BPAs of four criteria respectively}
\tiny
\begin{tabular*}{\columnwidth}{@{\extracolsep{\fill}}@{~~}llllll@{~~}}
\toprule
          &Performance           &$C1$                            &$C2$                            &$C3$                            &$C4$                                  \\
          &                      &$(\{IS\},\{NS\},\{IS,NS\})$     &$(\{IS\},\{NS\},\{IS,NS\})$     &$(\{IS\},\{NS\},\{IS,NS\})$     &$(\{IS\},\{NS\},\{IS,NS\})$    \\
\midrule
    DM1   &Weights               &[0.2857,0.5]                    &[0.4286,0.7857]                 &[0.0714,0.4286]                 &[0.3571,0.7143]\\
    \\
          &Supplier1             &([0.1714,0.3],[0.0571,          &([0.2755,0.5051],[0.0306,       &([0.0428,0.2572],[0.0143,       &([0.2143,0.4286],[0.0714,\\
          &                      &0.1],[0.7715,0.6])              &0.0561],[0.6939,0.4388])        &0.0857],[0.9429,0.6571])        &0.1429],[0.7143,0.4285])\\
          &Supplier2             &([0.1714,0.3],[0.0571,          &([0.2755,0.5051],[0.0306,       &([0.0357,0.2143],[0.0357,       &([0.1785,0.3572],[0.1785, \\
          &                      &0.1],[0.7715,0.6])              &0.0561],[0.6939,0.4388])        &0.2143],[0.9286,0.5714])        &0.3572],[0.6430,0.2856]) \\
          &Supplier3             &([0.1429,0.25],[0.1429,         &([0.1429,0.3928],[,0.1429,      &([0.0428,0.2572],[0.0143,       &([0.2381,0.4762],[0,\\
          &                      &0.25],[0.7142,0.5])             &0.3928],[0.7242,0.2144])        &0.0857],[0.9429,0.6571])        &0],[0.7619,0.5238])\\
          &Supplier4             &([0.1905,0.3333],[0,            &([0.2857,0.5238],[0,            &([0.0357,0.2143],[0.0357,       &([0.1785,0.3572],[0.1785, \\
          &                      &0],[0.8095,0.6667])¡¡¡¡¡¡¡¡¡¡¡¡ &0],[0.7143,0.4762])             &0.2143],[0.9286,0.5714])        &0.3572],[0,0])\\
          &Supplier5             &([0,0],[0.1905,                 &([0,0],[0.2857,                 &([0,0],[0.0476,                 &([0,0],[0.2381,\\
          &                      &0.3333],[0.8095,0.6667])        &0.5238],[0.7143,0.4762])        &0.2857],[0.9524,0.7143])        &0.4762],[0.7619,0.5238])\\
          &Supplier6             &([0.0571 0.1],[0.0714,          &([0.0306 0.0561],[0.2755,       &([0.0476 0.2857],[0,            &([0 0],[0.2381,\\
          &                      &0.3],[0.7715,0.6])¡¡¡¡¡¡¡¡¡¡¡¡  &0.5051],[0.6939,0.4388])        &0],[0.9524,0.7143])             &0.4762],[0.7619,0.5238])\\
                                                                   \\
    DM2   &Weights               &[0.3571,0.6428]                 &[0.2857,0.7857]                 &[0.0714,0.4286]                 &[0.2857,0.8571]                        \\
    \\
          &Supplier1             &([0.2143,0.3857],[0.0714,       &([0.1837,0.5051],[0.0204,       &([0.0357,0.2143],[0.0357,       &([0.1714,0.5143],[0.0571,\\
          &                      &0.1286],[0.7143,0.4857])        &0.0561],[0.7957,0.4388])        &0.2143],[0.9286,0.5714])        &0.1714],[0.7715,0.3143]) \\
          &Supplier2             &([0.2143 0.3857],[0.0714,       &([0.1837 0.5051],[0.0204,       &([0 0],[0.0476,                 &([0.1429 0.4285],[0.1429, \\
          &                      &0.1286],[0.7143,0.4857])        &0.0561],[0.7957,0.4388])        &0.2857],[0.9524,0.7143])        &0.4285],[0.7142,0.1430]) \\
          &Supplier3             &([0.1785,0.3214],[0.1785,       &([0.1429,0.3928],[0.1429,       &([0.0476,0.2857],[0,            &([0.1905,0.5714],[0,\\
          &                      &0.3214],[0.6430,0.3572])        &0.3928],[0.7142,0.2144])        &0],[0.9524,0.7143])             &0],[0.8095,0.4286])\\
          &Supplier4             &([0.2381,0.4286],[0,            &([0.1905,0.5238],[0,            &([0.0357,0.2143],[0.0357,       &([0.1429,0.4285],[0.1429,  \\
          &                      &0],[0.7619,0.5714])             &0],[0.8095,0.4762])             &0.2143],[0.9286,0.5714])        &0.4285],[0.7142,0.1430])\\
          &Supplier5             &([0,0],[0.2381,                 &([0,0],[0.1905,                 &([0,0],[0.0476,                 &([0.0571,0.1714],[0.1714,\\
          &                      &0.4286],[0.7619,0.5714])        &0.5238],[0.8095,0.4762])        &0.2857],[0.9524,0.7143])        &0.5143],[0.7715,0.3143])\\
          &Supplier6             &([0.0714,0.1286],[0.2143,       &([0.0204,0.0561],[0.1837,       &([0.0476,0.2857],[0,            &([0,0],[0.1905,\\
          &                      &0.3857],[0.7143,0.4857])        &0.5031],[0.7957,0.4388])        &0],[0.9524,0.7143])             &0.5714],[0.8095,0.4286]) \\
          \\
     DM3  &Weights               &[0.2857,0.7857]                 &[0.2857,1]                      &[0.1429,0.5714]                 &[0.2857,0.8571]\\
     \\
          &Supplier1             &([0.1714,0.4714],[0.0571,       &([0.1837,0.6429],[0.0204,       &([0.0817,0.3265],[0.0408,       &([0.1905,0.5714],[0,\\
          &                      &0.1571],[0.7715 0.3715])        &0.0714],[0.7957,0.2857])        &0.1632],[0.8775 0.5103])        &0],[0.8095,0.4286])\\
          &Supplier2             &([0.1714,0.4717],[0.0571,       &([0.1837,0.6429],[0.0204,       &([0.0952,0.3810],[0,            &([0.0816,0.2449],[0.1632,\\
          &                      &0.1571],[0.7715,0.3715])        &0.0714],[0.7957,0.2857])        &0],[0.9048,0.6190])             &0.4897],[0.7552,0.2654])\\
          &Supplier3             &([0.1429,0.3928],[0.1429,       &([0.1429 0.5],[0.1429,          &([0.0952,0.3810],[0,            &([0.1905,0.5714],[0,\\
          &                      &0.3928],[0.7142 0.2144])        &0.5],[7142,0])                  &0],[0.9048,0.6190])             &0],[0.8095,0.4286])\\
          &Supplier4             &([0.1905,0.5238],[0,            &([0.1905,0.6667],[0,            &([0.0817,0.3265],[0.0408,       &([0.1905,0.5714],[0,\\
          &                      &0],[0.8095 0.4762])             &0],[0.8095,0.3333])             &0.1632],[0.8775 0.5103])¡¡    ¡¡&0],[0.8095,0.4286])\\
          &Supplier5             &([0,0],[0.1905,                 &([0,0],[,0.1905,                &([0,0],[0.0952,                 &([0.0816,0.2449],[0.1632, \\
          &                      &0.5238],[0.8095 0.4762])        &0.6667],[0.8095,0.3333])        &0.3810],[0.9048 0.6190])        &0.4897],[0.7552,0.2654])\\
          &Supplier6             &([0.0571,0.1571],[0.1714,       &([0.0204,0.0714],[0.1837,       &([0.0952,0.3810],[0,            &([0,0],[0.1905,\\
          &                      &0.4714],[0.7715 0.3715])        &0.1429],[0.7959,0.2857])        &0],[0.9048 0.6190])             &0.5714],[0.8095,0.4286])   \\
\bottomrule
\end{tabular*}
}
\end{sidewaystable}

\begin{figure}[!h]
\centering
\includegraphics[scale=0.8]{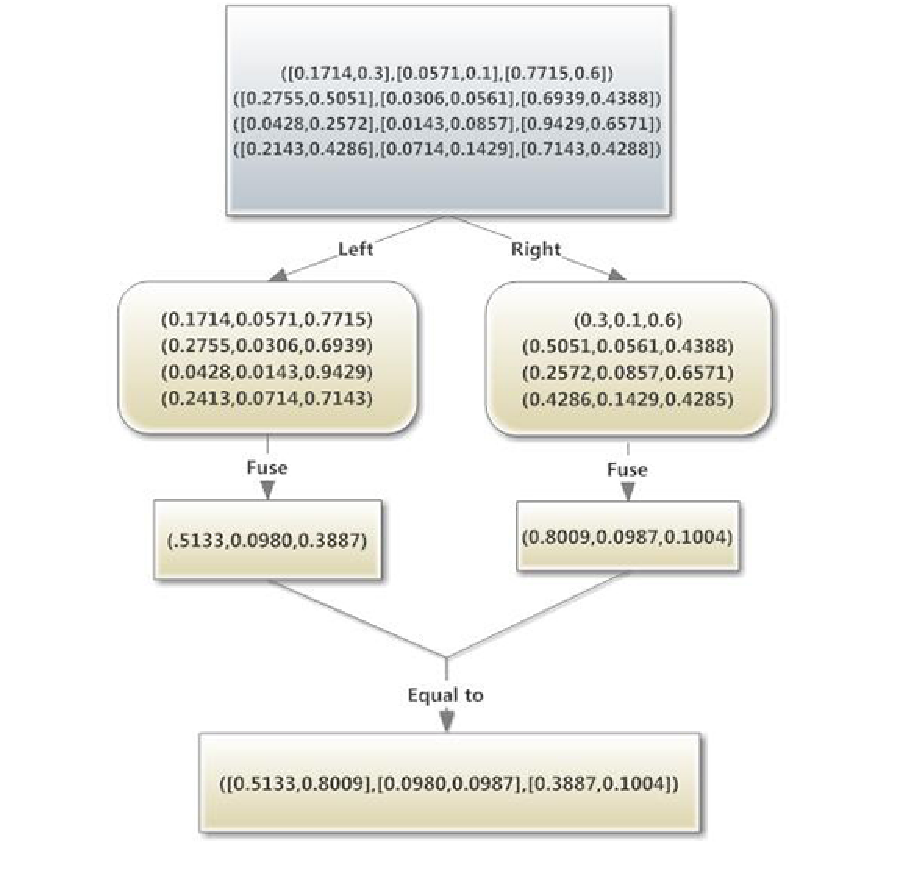}
 \caption{Combine the interval BPA}\label{xiaoliucheng}
\end{figure}

Now all the preparation before fusion is completed. Takes decision maker1's evaluation to supplier1 as an example to illustrate the procedure of combining the interval BPA(Figure \ref{xiaoliucheng}).

As the flow chart reveals, an interval BPA is equal to two classical BPAs groups which consists of the left part and the right part of the interval respectively. Then the four BPAs consisting of the left part are fused together based on DST, so do the other groups. The newly obtained two BPAs are of the classical properties and act as the left part and the right part of the new interval BPA respectively. In the same way, we can get all the interval BPAs which represent the comprehensive opinions of each supplier from each expert(Table \ref{Fuse interval data to get comprehensive information}).

\begin{table}[!h]
{\footnotesize
\caption{Fuse interval data to get comprehensive information.}\label{Fuse interval data to get comprehensive information}
\begin{tabular*}{\columnwidth}{@{\extracolsep{\fill}}@{~~}llllllll@{~~}}
\toprule
                  &Performance           & The left part of interval BPA             & The right part of interval BPA                                 \\
                  &                      & $(\{IS\},\{NS\},\{IS,NS\})$               & $(\{IS\},\{NS\},\{IS,NS\})$                                    \\
\midrule                                                                          \\
    DM1           &Supplier1             &(0.5133, 0.0980, 0.3887)                   &(0.8009, 0.0987, 0.1004)               \\
                  &Supplier2             &(0.4596, 0.1772, 0.3632)                   &(0.6881, 0.2353, 0.0766)               \\
                  &Supplier3             &(0.4003, 0.1895, 0.4102)                   &(0.6817, 0.2520, 0.0663)                        \\
                  &Supplier4             &(0.4938, 0.1246, 0.3815)                   &(0.7447, 0.1782, 0.0771)                         \\
                  &Supplier5             &(     0, 0.5804, 0.4196)                   &(     0, 0.8812, 0.1188)                         \\
                  &Supplier6             &(0.0734, 0.5131, 0.4135)                   &(0.1203, 0.7369, 0.1428)               \\
                                                                                                                                                   \\
                                                                            \\
    DM2           &Supplier1             &(0.4502, 0.1128, 0.4370)                   &(0.8108, 0.1268, 0.0624)               \\
                  &Supplier2             &(0.3929, 0.1800, 0.4271)                   &(0.6438, 0.3116, 0.0446)               \\
                  &Supplier3             &(0.3920, 0.2114, 0.3966)                   &(0.7549, 0.1995, 0.0456)                         \\
                  &Supplier4             &(0.4502, 0.1086, 0.4412)                   &(0.7774, 0.1821, 0.0405)                          \\
                  &Supplier5             &(0.0343, 0.5015, 0.4642)                   &(0.0387, 0.8905, 0.0708)               \\
                  &Supplier6             &(0.0854, 0.4518, 0.4628)                   &(0.0996, 0.7475, 0.1529)               \\                                                 \\
                                                                                                                      \\
    DM3           &Supplier1             &(0.4722, 0.0699, 0.4579)                   &(0.9206, 0.0456, 0.0338)               \\
                  &Supplier2             &(0.4015, 0.1829, 0.4155)                   &(0.8124, 0.1506, 0.0370)               \\
                  &Supplier3             &(0.4015, 0.1829, 0.4155)                   &(0.7903, 0.2097, 0)                               \\
                  &Supplier4             &(0.5034, 0.0221, 0.4746)                   &(0.9460, 0.0131, 0.0409)                         \\
                  &Supplier5             &(0.0500, 0.4868, 0.4631)                   &(0.0309, 0.9357, 0.0334)                         \\
                  &Supplier6             &(0.1051, 0.4620, 0.4833)                   &(0.0982, 0.8494, 0.0524)               \\
\bottomrule
\end{tabular*}
}
\end{table}

\textbf{Step 3.} Convert the decision makers' weights including crisp data and linguistic items into an interval number. And then discount the interval BPA of combined performance (obtained in Step 2) using the interval data to generate the interval BPA of each performance. Combine the interval BPA of all decision makers' to get the performance of each alternative.

Step 3. is similar to Step 2. In other words, the process of Step 3 is nearly the same as the last step essentially. It provides each interval BPA with another chance of applying Dempster-Shafer combination rule to be fused together. As a result, the BPA of ideal solution can be increased or reduced owing to the proporty of DST, which will contribute to our decision making greatly. Firstly, Using the same method(using Eq.(\ref{Eq13})), the new interval weights of decision makers' reliability can also be obtained as follows:
\[{W_{DM1}} = {{[\begin{array}{*{20}{c}}
{0.20},&{0.45}
\end{array}]} \mathord{\left/
 {\vphantom {{[\begin{array}{*{20}{c}}
{0.20},&{0.45}
\end{array}]} {0.95}}} \right.
 \kern-\nulldelimiterspace} {0.95}} = \left[ {\begin{array}{*{20}{c}}
{0.2105},&{0.4739}
\end{array}} \right]\]
\[{W_{DM2}} = {{[\begin{array}{*{20}{c}}
{0.35},&{0.55}
\end{array}]} \mathord{\left/
 {\vphantom {{[\begin{array}{*{20}{c}}
{0.35},&{0.55}
\end{array}]} {0.95}}} \right.
 \kern-\nulldelimiterspace} {0.95}} = \left[ {\begin{array}{*{20}{c}}
{0.3684},&{0.5789}
\end{array}} \right]\]
\[{W_{DM3}} = {{[\begin{array}{*{20}{c}}
{0.70},&{0.95}
\end{array}]} \mathord{\left/
 {\vphantom {{[\begin{array}{*{20}{c}}
{0.70},&{0.95}
\end{array}]} {0.95}}} \right.
 \kern-\nulldelimiterspace} {0.95}} = \left[ {\begin{array}{*{20}{c}}
{0.7368},&1
\end{array}} \right]\]

Then, we will use Eq.(\ref{Eq18}) to get the new interval BPAs. Next those interval BPAs will be fused like Step 2. Still take supplier1 as an example and get the result in Table \ref{Fuse the three DMs' evaluation of supplier1.}.

\begin{table}[!h]
{\footnotesize
\caption{Fuse the three DMs' evaluation of supplier1.}\label{Fuse the three DMs' evaluation of supplier1.}
\begin{tabular*}{\columnwidth}{@{\extracolsep{\fill}}@{~~}ccccc@{~~}}
\toprule
                  &                      & The left part of interval BPA             & The right part of interval BPA                                 \\
                  &                      & $(\{IS\},\{NS\},\{IS,NS\})$               & $(\{IS\},\{NS\},\{IS,NS\})$                                    \\
\midrule
    DM1           &                      &(0.1080, 0.0206, 0.8714)                   &(0.3795, 0.0468, 0.5737)                \\
                                                                            \\
    DM2           &                      &(0.1659, 0.0416, 0.7925)                   &(0.4694, 0.0734, 0.4572)                                                \\
                                                                                                                      \\
    DM3           &                      &(0.3479, 0.0515, 0.6006)                   &(0.9206, 0.0456, 0.0338)    \\
\midrule
\\
Fusion result     &                      &(0.4950, 0.0733, 0.4317)                   &(0.8849, 0.1017, 0.0135)    \\
\bottomrule
\end{tabular*}
}
\end{table}

Table \ref{Fuse the three DMs' evaluation of supplier1.}  reveals the the three evaluation of supplier1 offered by three experts. After the fusion, we get the final interval BPA which represents the most overall information about suppiler1.
Using the same method, all the suppliers' final interval BPA are obtained(Table \ref{Fuse the three DMs' evaluation of each supplier}). The rest steps are to compare these interval BPAs and rank the order to make decision.
\begin{table}[!h]
{\footnotesize
\caption{Fuse the three DMs' evaluation of each supplier.}\label{Fuse the three DMs' evaluation of each supplier}
\begin{tabular*}{\columnwidth}{@{\extracolsep{\fill}}@{~~}llllllll@{~~}}
\toprule
       Performance           & The left part of interval BPA           & The right part of interval BPA                                 \\
                             & $(\{IS\},\{NS\},\{IS,NS\})$             & $(\{IS\},\{NS\},\{IS,NS\})$                                    \\
\midrule
       Supplier1             & (0.4950, 0.0733, 0.4317£©               & (0.9696, 0.0201, 0.0103)\\
    \\
       Supplier2             & (0.4106, 0.1516, 0.4378)                & (0.8849, 0.1017, 0.0135) \\
    \\
       Supplier3             & (0.4019, 0.1878, 0.4104)                & (0.8851, 0.1149, 0)\\
    \\
       Supplier4             & (0.5135, 0.0485, 0.4380)                & (0.9765, 0.0110, 0.0125)\\
    \\
       Supplier5             & (0.0336, 0.5337, 0.4328)                & (0.0096, 0.9811, 0.0094)\\
    \\
       Supplier6             & (0.0845, 0.4895, 0.4260)                & (0.0473, 0.9339, 0.0189)    \\
\bottomrule
\end{tabular*}
}
\end{table}

\textbf{Step 4.} Combine the the lower part and the upper part of the interval BPA to get the final performance of each alternative.

The greatest advantage of using interval data is that it can retain the original information about the performance as much as possible during the fusion process. When it comes to rank the order of suppliers, a classical BPA consisting of crisp number seems to be a more effective means. Hence, we will combine the two parts of the interval BPA into one classical BPA in order to select the best supplier.

\textbf{Step 5.} Determine the final ranking order based on pignistic probability transformation(PPT).

The BPA of discernment ($m\left\{ {IS,NS} \right\}$) has some effects on the accuracy of making the best decision. To eliminate it, pignistic probability transformation is applied in our method. With the equation as
\begin{equation}
{Be{t_n}\left\{ {IS} \right\} = {m_n}\left\{ {IS} \right\} + {{{m_n}\left\{ {IS,NS} \right\}} \mathord{\left/
 {\vphantom {{{m_n}\left\{ {IS,NS} \right\}} 2}} \right.
 \kern-\nulldelimiterspace} 2}}
 \end{equation}
 the final belief degrees of each supplier are showed in Table \ref{Transfer the interval BPA back to classical BPA}. And according to these data, the order is easily ranked as supplier 4 $succ$ supplier 1 $succ$ supplier 2 $succ$ supplier 3 $succ$ supplier 6 $succ$ supplier 5. Apparently, supplier 4 is the best selection. It coincides with the results presented in paper\cite{Deng2011A}. Furthermore, the final rank order is also coincided with the the left part or the right part of interval BPA, which proves the feasibility and validity of our new method.
 \begin{table}[!h]
{\footnotesize
\caption{Convert the interval BPA back to classical BPA}\label{Transfer the interval BPA back to classical BPA}
\begin{tabular*}{\columnwidth}{@{\extracolsep{\fill}}@{~~}lllc@{~~}}
\toprule
       Performance           & Fused results                 & bet(IS)       &Final ranking order                             \\
\midrule
       Supplier1             & (0.9833, 0.0119, 0.0048£©     & 0.9857              &2 \\
    \\
       Supplier2             & (0.9177, 0.0752, 0.0072)      & 0.9213              &3 \\
    \\
       Supplier3             & (0.9129, 0.0873, 0)           & 0.9129              &4 \\
    \\
      Supplier4              & (0.9879, 0.0063, 0.0058)      & 0.9908              &1 \\
    \\
      Supplier5              & (0.0050, 0.9910, 0.0042)      & 0.0071              &6 \\
    \\
      Supplier6              & (0.0287, 0.9625, 0.0090)      & 0.0332              &5 \\
\bottomrule
\end{tabular*}
}
\end{table}

 \section{Conclusion}\label{Conclusion}
In reality, MCDM problem faces a mass of fuzzy information inevitably. To handle this problem, a new method is proposed based on interval data fusion. The fuzzy data is collected in the form of interval data in our method. Compared with the original method, our method has remarkable superiority in dealing with the fuzzy information. A supplier selection example is used to illustrate the detailed procedures of our method and the result proves its correctness adequately. Our new method is worthy being taken into consideration when the fuzzy data grow rapidly as the system develops. Furthermore, our method holds quantities of opportunities to apply, especially in the fields like social, economy and so on.

\section{Acknowledgement}
The work is partially supported by National Natural Science Foundation of China (Grant No. 61671384), Natural Science Basic Research Plan in Shaanxi Province of China (Program No. 2016JM6018), Aviation Science Foundation (Program No. 20165553036), the Fund of SAST (Program No. SAST2016083).

\bibliographystyle{elsarticle-num-names}
\bibliography{myreference}

\end{document}